\documentclass[10pt,twocolumn,letterpaper]{article}

\usepackage{wacv}
\usepackage{times}
\usepackage{epsfig}
\usepackage{graphicx}
\usepackage{amsmath}
\usepackage{amssymb}
\usepackage{booktabs}
\usepackage[accsupp]{axessibility}

\usepackage{makecell}

\def\wacvPaperID{1404} % Enter the WACV Paper ID here
\wacvalgorithmstrack

\wacvfinalcopy % *** Uncomment this line for the final submission

\ifwacvfinal
\usepackage[breaklinks=true,bookmarks=false]{hyperref}
\else
\usepackage[pagebackref=true,breaklinks=true,colorlinks,bookmarks=false]{hyperref}
\fi

\begin{document}

%%%%%%%%% TITLE
\title{3D-LatentMapper: View Agnostic Single-View Reconstruction of 3D Shapes}

\author{Alara Dirik$^{1,2}$ \quad Pinar Yanardag$^{2}$\\
$^1$Hugging Face, $^2$Bogazici University \\
{\tt\small alara@huggingface.co, pinar.yanardag@boun.edu.tr}
}

\maketitle
%\thispagestyle{empty}

%%%%%%%%% ABSTRACT
\begin{abstract}
   Computer graphics, 3D computer vision and robotics communities have produced multiple approaches to represent and generate 3D shapes, as well as a vast number of use cases. However, single-view reconstruction remains a challenging topic that can unlock various interesting use cases such as interactive design. In this work, we propose a novel framework that leverages the intermediate latent spaces of Vision Transformer (ViT) and a joint image-text representational model, CLIP, for fast and efficient Single View Reconstruction (SVR). More specifically, we propose a novel mapping network architecture that learns a mapping between deep features extracted from ViT and CLIP, and the latent space of a base 3D generative model. Unlike previous work, our method enables view-agnostic reconstruction of 3D shapes, even in the presence of large occlusions. We use the ShapeNetV2 dataset and perform extensive experiments with comparisons to SOTA methods to demonstrate our method's effectiveness. 
\end{abstract}

%%%%%%%%% BODY TEXT
\section{Introduction}
Generative Adversarial Networks (GANs) \cite{NIPS2014_5423} are generative models that revolutionized computer vision. Today, GANs are widely used for various visual tasks due to their success in generating high-quality images. To name a few, image generation \cite{Karras2019ASG, Karras2020AnalyzingAI}, image manipulation  \cite{wang2017highresolution}, image denoising \cite{wang2019spatial,li2019single}, image super-resolution \cite{Sun_2020} and domain translation \cite{CycleGAN} are only some of the many creative uses of generative models. 

Despite the impressive capabilities of GANs, these advances are overwhelmingly limited to 2D image generation tasks whereas 3D generation and reconstruction from single-view images remain challenging. This is mainly because one of the greatest challenges to the development of 3D generative models is converging on the right representation as it is often not possible to apply image-based methods to 3D data. Whereas RGB images have become the quasi-standard in 2D vision tasks, common 3D representations such as point clouds, meshes and other compact surface representations all pose significant problems during training, which makes it difficult to agree on a common representation. For example, direct extension of 2D pixel grids to 3D voxel grids \cite{Wu2016LearningAP} to train convolutional networks significantly limits resolution due to the high memory demands of 3D convolutions. Mesh datasets, on the other hand, often require re-meshing of mesh objects to create an even number of faces and vertices across the dataset, which often leads to a loss in detail depending on the shape topologies and the re-meshing method used. Similarly, while point clouds are very popular for generative tasks \cite{Achlioptas2018LearningRA,cai2020learning,Li2021SPGANS3}, they are limited in their ability to model sharp features or high-resolution color textures. Hence, the quality, flexibility and fidelity of 3D encoding and generation approaches are limited by the representation used. To address these issues, several alternative representations such as NeRF \cite{mildenhall2020nerf} and DeepSDF \cite{Park2019DeepSDFLC}, as well as hybrid methods collectively known as neural implicit representations, have been proposed in the recent years to define implicit surface or volume representations using neural networks. 

In this work, we focus on the Single View Reconstruction (SVR) task and propose a novel framework that leverages large-scale pre-trained vision and joint vision-text models. Unlike previous works that directly operate on explicit 3D representations, our framework solely operates within the latent space of a base 3D generative model, Deep Implicit Templates (DIT) \cite{Zheng2021DeepIT}, to enable real-time reconstruction. More specifically, we propose a highly effective and fast method that leverages Vision Transformer (ViT) \cite{Dosovitskiy2021AnII} and CLIP \cite{Radford2021LearningTV}, a joint image-text representation model trained on a large unsupervised dataset, to map images to the latent space of DIT in a view-agnostic manner. While we use DIT as our base generative model, we note that our framework can be used with any 3D generative model. For all experiments, we use the \textit{car} and \textit{airplane} categories of the ShapeNetV2 dataset \cite{Chang2015ShapeNetAI} and compare our framework to two state-of-the-art competitor methods: AtlasNet \cite{groueix2018} and IM-NET \cite{Chen2019LearningIF}. We show through extensive experiments that our framework achieves fast and effective reconstruction without pose constraints. We list our main contributions as follows:

\begin{itemize}
    \item We create an augmented ShapeNet dataset with more accurate text descriptions and high-quality multi-view renders.
    \item We propose a novel, view-agnostic SVR network architecture that can map input images to the learned latent space of a 3D generative model, enabling image-based reconstruction and direct manipulation.
    \item We show that using text-conditioned image features yields semantically consistent reconstructions, even in cases where the target object is not included in the training dataset.
\end{itemize}

\section{Related Work}
\subsection{3D Shape Representations} 
\label{sec:reprclassic}
One of the most pressing research questions in the domain of 3D vision is how to best represent 3D data. Unlike the 2D vision domain, where RGB pixel representations have become the standard for research, 3D vision research still uses a wide variety of explicit (e.g. point clouds, voxels, meshes), implicit and hybrid shape representations. 

One of the most popular explicit shape representations is the point cloud, which represents shapes as a set of 3D coordinates in (x, y, z) format. As many sensors used in the industry, such as LIDAR and depth cameras, output point clouds, the point cloud format is extensively used in popular 3D vision problems such as reconstruction of 3D shapes \cite{park2011high}, 3D object classification \cite{fan2017point, qi2017pointnet}, and segmentation \cite{qi2017pointnet}. While popular, point clouds provides no information on how the points are connected, are order invariant and often yield noisy reconstructions and generations. Another very popular 3D representation is the mesh format, which describes each shape as a set of triangles, where each edge of each face is a connection between two vertices. While meshes are better suited to describe the topology, they pose significant problems such as the question of how to deal with shapes with unequal number of vertices and faces. Voxel format is yet another popular 3D representation format that describes objects as a fixed-sized volume occupancy matrix. While this dense grid structure is particularly suitable for CNN-based architectures, voxel format requires high-resolution in order to describe fine-grained surface descriptions and hence, cannot be generalized to articulate shapes.

Finally, there have been a large number of neural implicit representations proposed in the recent years that seek to overcome the shortcomings of the classical 3D shape representations. Neural implicit representations represent 3D shapes as learned functions that map 3D coordinates to a signed distance function (SDF) \cite{Park2019DeepSDFLC}, or a binary occupancy value \cite{Chen2019LearningIF,Mescheder2019OccupancyNL}. While SDF values denote how far a given point (x,y,z) is from the closest normal surface point, occupancy fields tell if the query point (x, y, z) is within the shape surface boundaries. Hence, both SDF based and occupancy based representations aim to create a lightweight and continuous shape representation that is infinitely scalable in resolution. Despite their significant advantages, a drawback of implicit representations is they require aggressive sampling and querying of 3D coordinates, followed by a ray-marching or sphere-tracing process in order to construct explicit surfaces. Neural volume rendering methods such as NeRF \cite{Mildenhall2020NeRFRS} are tangent to this problem as NeRF proposes a method to synthesize views with user specified camera intrinsics and perspective without explicitly constructing a surface. In practice, this is achieved by generating radiance fields along specified ray paths. 

\subsection{3D Shape Generation} 
\label{sec:3d-gen}
3D generative methods can be classified into two broad categories based on their main motivation: 3D representation learning and 3D shape generation.  While these tasks are often intertwined, 3D representation learning is arguably the most popular 3D vision task with a wide range of application domains such as 3D object detection, data compression and retrieval, navigation (e.g. SLAM) and symbol generation. 

Previous works on 3D representation learning use a wide array of shape representations and methods. PointNet \cite{fan2017point, qi2017pointnet}, for example, processes point cloud data and extracts global shape features via max-pooling operations for object classification. The shape features extracted by PointNet are also widely used as input for point and occupancy generation networks \cite{Chen2019LearningIF}. As described in Section \ref{sec:reprclassic}, an important limitation of working with point clouds is that they consist of unordered points and converting them to meshes often fail to produce consistent and watertight surfaces. Other works such as \cite{han2016mesh} propose using convolutional restricted Boltzmann machines for unsupervised learning of mesh features, which can be used for training linear and non-linear 3D object classifiers. AtlasNet \cite{groueix2018} proposes representing 3D shapes as a collection of parametric surface elements and tries to learn a mapping from 2D squares sampled from images or point clouds coupled with learned latent shape features to 3D points. However, their approach does not exploit locality information and results in lower fidelity generation outputs. 
Works such as \cite{Park2019DeepSDFLC, Zheng2021DeepIT, Mescheder2019OccupancyNL} propose conditional frameworks to learn implicit surface or volume functions while simultaneously learning an encoding for each training shape. Similarly, IM-GAN \cite{Chen2019LearningIF} proposes to use an implicit decoder, IM-NET, in conjunction with features learned by a latent-GAN model \cite{Achlioptas2018LearningRA} to yield a general 3D generative model. Because the outputs of this method and related methods are implicit in the network weights, high-resolution results typically cannot be visualized or exported in real time.
Other work, \cite{Sitzmann2020ImplicitNR, Takikawa2021NeuralGL} propose overfitting a neural implicit function per shape to encode and reconstruct objects as needed. Another branch of research focuses on learning continuous data distributions for novel shape synthesis. For example, SP-GAN \cite{Li2021SPGANS3} proposes an unsupervised sphere-guided generative model for directly synthesize point clouds. Others propose various CNN-based GAN architectures \cite{Achlioptas2018LearningRA,Hui2020ProgressivePC,Shu20193DPC} to generate novel shapes in the point cloud format.

\subsection{3D Shape Reconstruction} \label{sec:3d-recon}
An important line of research in 3D computer vision is 3D reconstruction from single or multiple 2D views, or partial 3D views. As 3D data is often acquired from LIDAR sensors in the form of points clouds, and thus suffer from issues such as occlusion, most work on 3D reconstruction focuses on reconstruction from single or multi-view partial point clouds. Works such as PoinTr++ \cite{Yu2021PoinTrDP} and IM-NET \cite{Chen2019LearningIF} operate on point clouds and employ iterative refinement schemes that seek to minimize the average Chamfer distance between the reconstructed point cloud and the ground truth. Moreover, SDF and occupancy-based representations have been successfully used for reconstruction and 3D shape completion from partial point clouds \cite{Mescheder2019OccupancyNL, Xu2019DISNDI, Chibane2020ImplicitFI}.

Another line of work seeks to reconstruct 3D shapes from single or multi-view images. For example, works such as \cite{kar2015category, choy20163d} leverage the grid-like architecture of the voxel format to train CNNs on single and multi 2D views respectively to reconstruct the 3D object in voxel format. Other work such as Pixel2point \cite{Afifi2021Pixel2point3O} and AtlasNet propose CNN-based architectures that directly map pixels of a single-view image to 3D points to create point cloud and mesh outputs respectively. Similarly, works such as DeepI2P \cite{Li2021DeepI2PIC} aim to register each pixel in a single-view to a point in a LIDAR acquired point cloud for SLAM purposes. In our work, we show that it is possible to perform higher quality reconstructions without a time-consuming optimization scheme by restricting the mapping process to the latent space of the generative model instead of operating on explicit 3D representations and directly mapping pixels to points on meshes or point clouds. For the sake of brevity, we exclude multi-view reconstruction from related work.

\section{Methodology} \label{sec:svr}
In this section, we describe our Single-View Reconstruction framework, which consists of multiple modules and is trained in 3 stages. Our framework, denoted as 3D-LatentMapper, essentially learns a mapping between dense features extracted from ViT and CLIP models pre-trained on large, general-purpose image datasets. Hence, we first describe our dataset generation process and preprocessing steps to extract features from  ViT and CLIP models. Then, we describe our framework and training procedure, as well as a real-to-synthetic image translation module that enables processing real images for reconstruction.

\subsection{Dataset Generation}
We train 3D-LatentMapper using sets of matching synthetic images and encoded 3D shapes for each shape category. Based on the size of the ShapeNetV2 categories, intra-category diversity of shapes and availability of real-world image datasets, we choose to use \textit{car} and \textit{airplane} categories and create synthetic datasets for these categories as follows. We first encode each 3D shape into DIT's latent space by initializing a latent vector and optimizing it via back-propagation. To this end, we randomly sample $10,000$ 3D coordinates from a unit sphere and forward propagate the sampled coordinates through the pre-trained DIT model to acquire their corresponding predicted SDF values. We then compute the L1 loss between the ground truth and predicted SDF values of the query coordinates, and optimize the latent vector via back-propagation. After converging to an optimized latent vector, we use the latent vector to compute the SDF values of randomly  sampled coordinates and acquire its corresponding mesh via ray-marching of SDF values. We save the latent vectors (encoded 3D shapes) along with the meshes for quantitative evaluation purposes. In order to train a view-agnostic SVR model, we render N=9 views of each generated mesh from a fixed camera view by rotating each mesh $\mathcal{M}$ by $0$, $30$, $60$, $90$, $120$, $150$, $180$, $210$ and $240$ degrees around the y-axis. Our reasoning in rendering generated meshes instead of ground truth meshes is generated meshes are better representatives of of DIT's learned latent space and are easier to learn from.  

In addition to synthetic image datasets, we use a publicly available car image dataset, the Car Connection Dataset, to train a real-to-synthetic image translation model as described in Section \ref{sec:real-to-synth}. The Car Connection Dataset is automatically scraped from various websites and  consists of 66,079 car images, including partial and detail views (e.g. car interior, texture details). In order to filter these irrelevant images, we use a set of engineered text templates that describe the target category (e.g. "Photo of a car") and compute the average cosine distance between the CLIP embeddings of each image and the set of text templates. A full list of the text templates used can be found in the Appendix \ref{app:templates}. This process is formulated in Equation \ref{eq:clip-dist}:
\begin{equation}
 \mathcal{D}_{\text{CLIP}} = \frac{\sum_{j=1}^N D_{\text{CLIP}}(\mathcal{I}, \mathcal{T}_{\text{j}})}{N}
 \label{eq:clip-dist}
\end{equation}

where $\mathcal{I}$ is an image from the dataset to be filtered, $\mathcal{T}_j$ is a text description of the target shape category embedded in a text template from a list of $N$ templates. Once the average CLIP distance of each image is computed, we use the mean ($\mu$) and the standard deviation ($\sigma$) of the computed distances to set a threshold such that images that yield an average distance of more than two standard deviations away from the mean are filtered out. The filtering process is formulated in Equation \ref{eq:filter}.

\begin{equation} 
  \mathcal{C}_{\mathcal{I}} =\begin{cases}
    0, & \text{if $D_{\text{CLIPavg}}(\mathcal{I}) > \mu + 2*\sigma $}.\\
    1, & \text{otherwise}.
  \end{cases}
  \label{eq:filter}
\end{equation}

\subsection{Single-View Reconstruction}
Next, we propose a novel SVR framework that leverages ViT and CLIP to map single-view images to the learned latent space of DIT. Our motivation is as follows: DeepSDF and DIT learn SDF functions and do not generate meshes. Instead, they rely on aggressive point sampling, SDF value retrieval and ray-marching to generate a mesh. This mesh generation process is not only time-consuming (90 seconds per generation) but also non-differentiable, meaning we cannot optimize the generated shapes directly. In our work, we aim to overcome this issue by directly predicting the latent code that corresponds to the input image without generating a mesh. 

Given a set of 3D shapes $(m_0, m_1, ..., m_N)$ and a DIT model $\mathcal{G}$ trained on a target shape category, let $\mathbf{c_i} \in \mathcal{R}^d$ denote a d-dimensional latent vector that corresponds to the $i$th 3D shape such that $0 \leq i \leq N$. Our goal is to map a single-view image $\mathcal{I}_i$ to a latent vector $\mathbf{c_i}$, such that querying DIT with $\mathbf{c_i}$ and sampled 3D coordinates and subsequently ray-marching the predicted SDF values reconstructs the corresponding 3D shape. To achieve this, we propose a three-stage mapping network architecture, which we denote as 3D-LatentMapper, and leverage ViT and CLIP as dense feature descriptors. We use CLIP and ViT instead of popular dense feature descriptors such as ResNet \cite{He2016DeepRL} trained on large datasets for two main reasons: (i) CLIP is a joint text-image representation model that can be used to extract semantically-grounded features, which in return are more likely to lead to contextually relevant reconstructions even in the case of previously unseen objects; (ii) The ViT architecture provides a rich latent space with positional information and has been show to be more robust to variations in scale and pose and less biased to textures and lighting. This provides a significant opportunity to use ViT as a dense feature descriptor.

\begin{figure*}[t]
	\centering
	\includegraphics[width=0.95\linewidth]{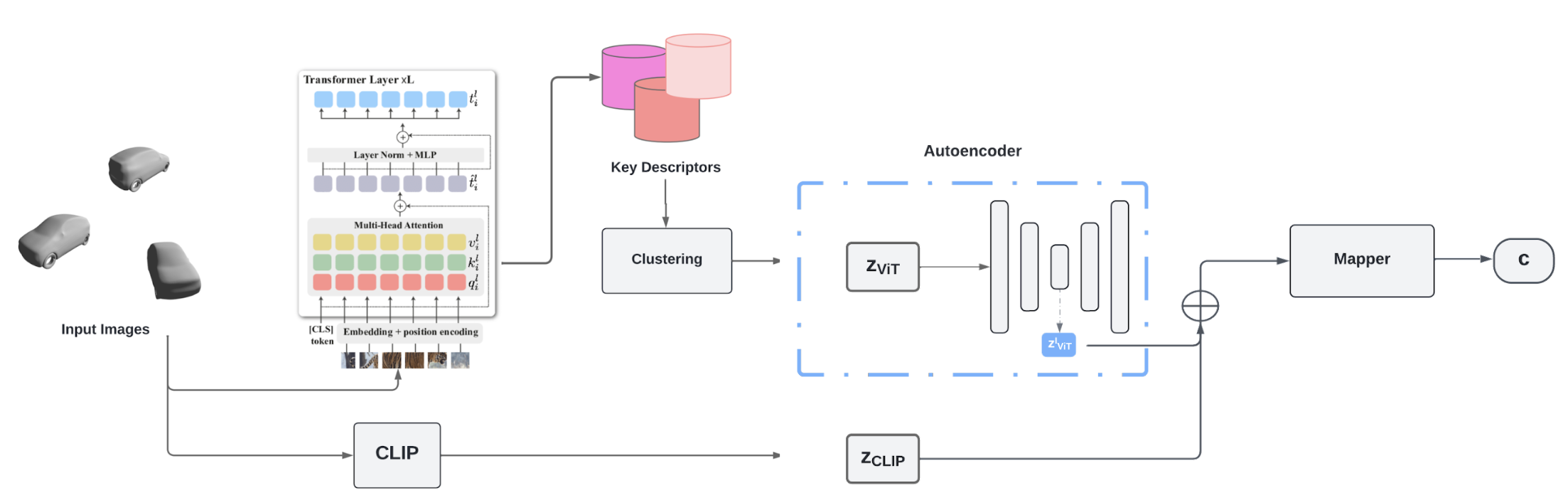}
	\caption{An overview of the 3D-LatentMapper framework: (i) Each image is fed into CLIP's image encoder to extract image embeddings and into ViT to extract key descriptors from all layers. The extracted key descriptors from $N=500$ random images are then clustered to assign descriptor labels and create 4235-dimensional descriptors. (ii) An autoencoder is trained to compress and reconstruct ViT descriptors. Once trained, the autoencoder's weights are frozen and the model is used to extract dense features from the bottleneck layer. (iii) Compressed ViT descriptors and CLIP embeddings are concatenated and used as input to a mapping module that maps the combined dense features to the latent space of DIT. }
	\label{fig:framework}
	\vspace{-1em}
\end{figure*}

For preprocessing, we use a pre-trained CLIP model's image encoder to extract 512-dimensional embeddings of the input images: $CLIP(\mathcal{I}) = \mathbf{z}_{CLIP}$ such that $\mathbf{z}_{CLIP} \in \mathcal{R}^{512}$. 

Additionally, we use a pre-trained ViT model and leverage the findings of previous works \cite{Caron2021EmergingPI} to extract dense features from images by forward propagating the input image $\mathcal{I}_i$ up until the target intermediate layer of ViT. Unlike CNN-based models, ViT represents each image as a collection of $n$ patches, where each image patch $p_i$ is directly associated with its own query, key, value, and token in each layer $l$: $\{q_i^l, k_i^l, v_i^l, t_i^l\}$. Similarly, whereas earlier layers feature coarse features such as edges or local texture patterns and deeper layers capture higher level concepts in CNN-based models, it has been shown \cite{Amir2021DeepVF} that ViTs feature a different type of representation hierarchy.  More precisely, the latent spaces of the shallow layers mainly capture the position of the input image patch, while in the deeper layers this position bias is reduced in favor of semantic features. 

Leveraging the findings of previous work on ViT's latent spaces \cite{Amir2021DeepVF}, we extract the key values $k_i^l$ across  all layers and all images to use as dense ViT descriptors, as they have been shown to be less sensitive to background noise in images. As the extracted keys are across multiple layers, it is not possible to perform clustering using all descriptors due to memory constraints. To overcome this issue: (i) we extract all key descriptors from $N=500$ randomly sampled synthetic images and additionally sub-sample 20\% of the extracted descriptors; (ii) we adopt a bag-of-descriptors approach and cluster the descriptors using the K-means algorithms with the number of clusters set to $K=20$. We note that clustering the descriptors of an image $\mathcal{I}_i$ yields a dense feature vector $\mathbf{z}_{ViT}$ such that $\mathbf{z}_{ViT} \in \mathcal{R}^{4235}$. An overview of our preprocessing pipeline is shown in Figure \ref{fig:framework}.

After extracting descriptors from synthetic images, we successively train two additional modules to map the dense features of each image to the latent space of DIT. As shown in Figure \ref{fig:framework}, we first use the dense features extracted from ViT, $\mathbf{z}_{ViT}$, to train an feed-forward autoencoder model. The autoencoder's encoder and decoder each consist of a single hidden layer and is trained for 100 epochs. By training an autoencoder, we seek to further compress the dense ViT features to a 512-dimensional space such that, after the autoencoder is trained, we freeze the model weights and use it to extract latent vectors from the autoencoder's bottleneck. We denote the latent vectors extracted from the bottleneck layer as $\mathbf{\bar{z}}_{ViT}$ and concatenate them CLIP embeddings to create a combined descriptor $\mathbf{z} \in \mathcal{R}^{1024}$. Finally, we use the concatenated dense feature descriptors as input and corresponding DIT latent vectors $c$ as output into a feed-forward network with three hidden layers followed by hyperbolic tangent activations. We train this model using an L1 loss and the Adam optimizer for 500 epochs. Once trained, our model learns a mapping $z \rightarrow c$, such that the predicted latent vector can be directly fed into DIT to generate a mesh or can be manipulated using unsupervised and supervised manipulation methods. 

\subsection{Real-to-Synthetic Image Translation} \label{sec:real-to-synth}
While our framework is highly effective at reconstructing 3D shapes from single-view synthetic images, we propose an additional real-to-synthetic image translation module to enable reconstruction from real images. To this end, we use CycleGAN \cite{CycleGAN} - an image translation framework that consists of two generators and two discriminators trained simultaneously. More specifically, CycleGAN learns a bidirectional mapping between two unaligned image datasets, $X$ and $Y$. The two generators  $G: X \rightarrow Y$ and $F: Y \rightarrow X$ are trained jointly with discriminators $D_{x}, D_{y}$ that encourage the generation of realistic images in the corresponding domain. The model is trained with cycle consistency loss terms $F(G(x)) \approx x, G(F(y)) \approx y$ for $x \in X, y \in Y$ in addition to the classical GAN losses of $G$ and $F$, the composite loss is given in Equation \ref{eq:cyclegan}.
\begin{equation} \label{eq:cyclegan}
    \begin{aligned}
    \mathcal{L}_{\text {CycleGAN }} \left( G, F, D_{x}, D_{y} \right) &= \mathcal{L}_{\text {GAN }} \left( G, D_{Y}, X, Y \right )\\
    & + \mathcal{}{L}_{\text {GAN }} \left( F, D_{X}, Y, X \right)\\
    & + \lambda_{\text {cycle }} \mathcal{L}_{\text {cycle }}(G, F)
    \end{aligned} 
\end{equation} 

 We note that our main goal is to leverage CycleGAN's lack of need for aligned image datasets for training. Once trained, we find that CycleGAN is highly effective at translating real images to synthetic images.

\section{Experiments}
\label{chapter:experiments-and-results}
\subsection{Datasets}
For the quantitative comparisons, we use the car and airplane categories of the ShapeNetV2 dataset \cite{Chang2015ShapeNetAI}, consisting of 3509 and 4045 shapes respectively. We choose these two categories as they contain sufficiently structurally distinct shapes and demonstrate the range of our method. For Real-to-Synthetic image translation, we use a publicly available dataset - the Car Connection Dataset, which consists of 66,079 car images. The Car Connection Dataset is automatically scraped from various websites and includes partial car images and detail views (e.g. interior, texture details). As described in Section \ref{sec:svr}, we use CLIP to filter the dataset and create a dataset of 41,122 car images.

\subsection{Evaluation Metrics}
\label{sec:exp:metrics}
 For our SVR experiments, we compare the generated shapes to ground truth shapes using Chamfer distance (CD) and Earth Mover's Distance (EMD) metrics. CD and EMD metrics are computed by sampling 5000 points from the generated and ground truth shapes, matching the nearest points across the sets of points and computing their sum of Euclidean and Manhattan distances respectively. For reconstruction from single-view real images, we perform qualitative and quantitative evaluation and compare our method to state-of-the-art SVR methods. Finally, we report the average inference time of our method and competing methods as real-time manipulation and reconstruction is critical for interactive applications, such as shape exploration and design.

\begin{table*}[t]
  \centering
  \newcommand{\qualimg}[1]{\raisebox{-.5\height}{\includegraphics[width=0.15\linewidth]{#1}}}
  \begin{tabular}{m{1cm}cccccc}
  Input  &
  \qualimg{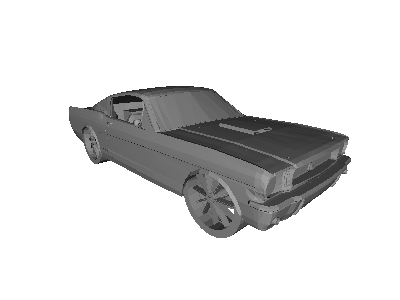} & 
  \qualimg{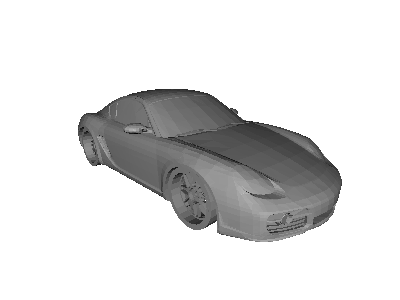} & 
  \qualimg{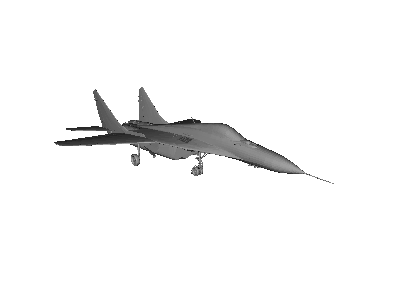} &  
  \qualimg{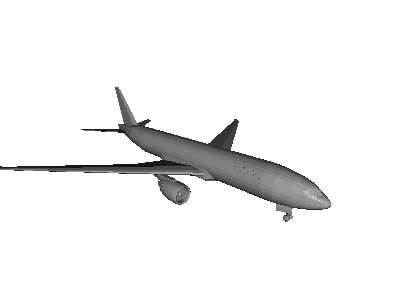}  \\
  \hline
  AtlasNet-Sphere &
  \qualimg{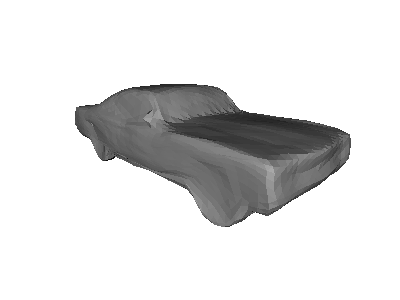} & 
  \qualimg{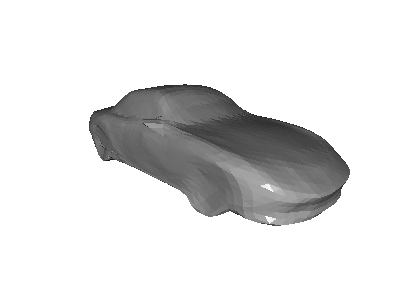} & 
  \qualimg{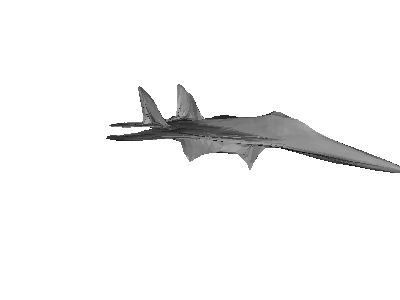} & 
  \qualimg{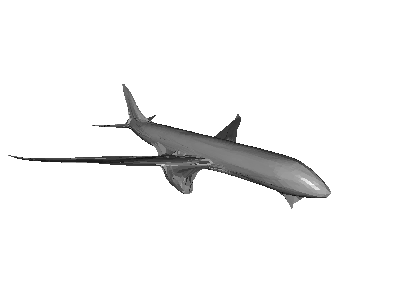} \\
  AtlasNet-25 &
  \qualimg{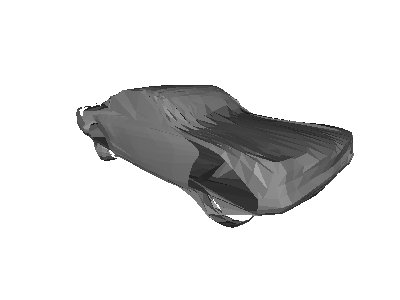} & 
  \qualimg{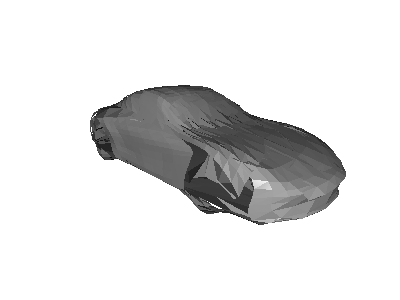} & 
  \qualimg{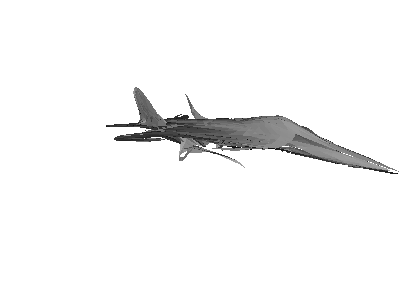} & 
  \qualimg{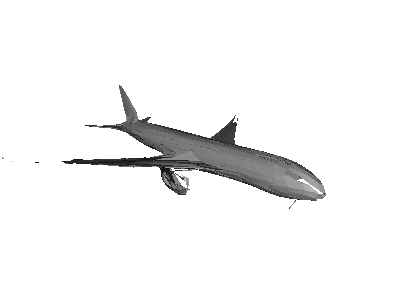} \\
  IM-Net &
  \qualimg{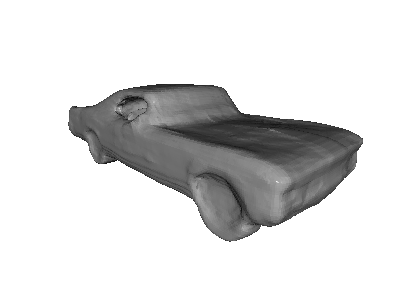} & 
  \qualimg{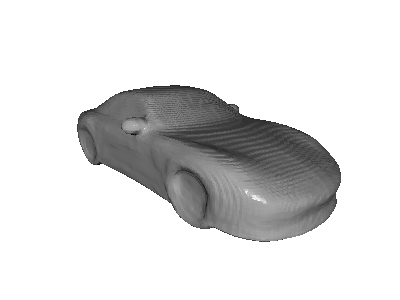} & 
  \qualimg{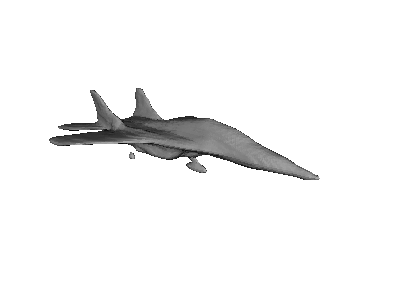} & 
  \qualimg{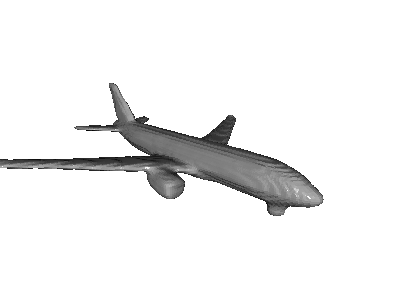}    \\
  \hline
  \makecell[l]{\begin{tabular}{@{}l@{}}Ours \\ \end{tabular}} &
  \qualimg{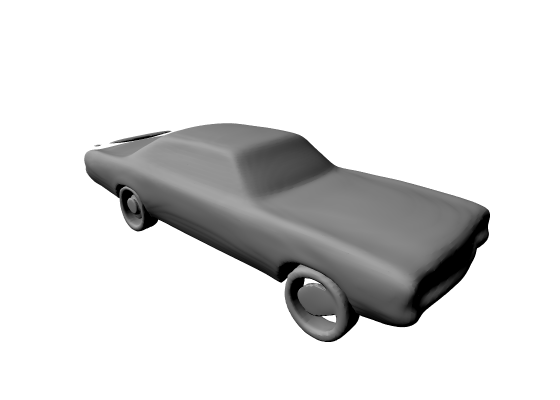} & 
  \qualimg{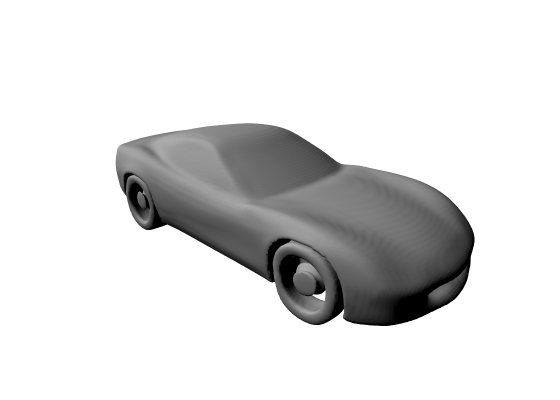} & 
  \qualimg{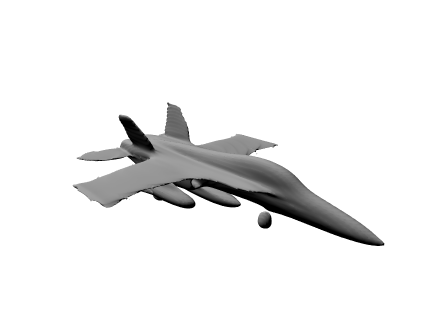} &  
  \qualimg{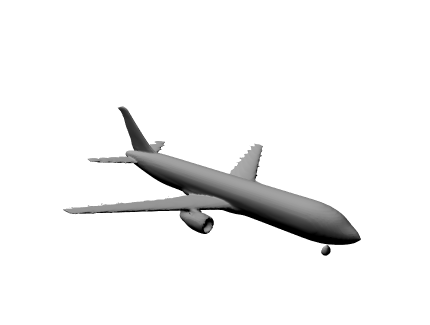}   \\

  \end{tabular}
  \caption{Qualitative reconstruction results: note that our method achieves the highest-fidelity 3D reconstructions and runs in real-time. Input images for SVR are shown in top row.}
  \label{tb:qual_svr2}
  \vspace{-1em}
\end{table*}

\subsection{Experimental Setup} \label{sec:exp:setup}
For all experiments, we train generative and reconstruction models on each shape category separately. We use the ShapeNetV2 dataset and create training, validation and test sets with a split ratio of 80-10-10\% for each shape category. We use the official implementation of Deep Implicit Templates and train models from scratch for the \textit{Car} and \textit{Airplane} categories. To train DIT, we create an improved preprocessing pipeline to eliminate non-watertight meshes, implement a new point sampling strategy to increasingly aggressive sampling near the surface, set the latent vector size the $512$ and keep the rest of hyperparameters the same. To encode ground truth 3D meshes into the latent space of DIT, we use a fixed 500-step optimization process and use the Adam optimizer with the default hyperparameters. 

To extract CLIP features, we use the official implementation of CLIP \cite{Radford2021LearningTV} with a ResNet \cite{He2016DeepRL} backbone to extract dense features and train the image encoder module. Similarly, we use the official implementation of Dino-ViT \cite{Amir2021DeepVF} to extract key descriptors from input images. Due to memory constraints, we use Scikit-Learn's Mini Batch K-means implementation to cluster the extracted key descriptors. We apply random zoom-in/out and rotation transformations to all images to improve the generalization capabilities of our method to real images. For the autoencoder module of our SVR framework, we use L2 loss and the Adam optimizer with default hyperparameters and train the model for 100 epochs for each category. For the mapper module of our SVR framework, we use an L1 loss and the Adam optimizer with default hyperparameters and train the model for 1000 epochs with early stopping enabled after 5 epochs to avoid overfitting. To map real images to synthetic images, we  train a bi-directional image-to-image translation model separately for the car category using the official implementation of CycleGAN \cite{Rai2018UnpairedIT}, and train the model on unaligned pairs of real images of the target category and rendered images of encoded 3D shapes. We train the CycleGAN model for 500 epochs with batch size of 8, using the default hyperparameters. 

For comparisons with other 3D generation and SVR methods, we use the official PyTorch implementations of AtlasNet and IM-NET with the default hyperparameters. Additional details of single-view reconstruction experiments with AtlasNet and IM-NET are provided in Appendix \ref{app:svr-comp}. We run all experiments, including model training and inference on a single NVIDIA TitanRX GPU.

\section{Single-View Reconstruction}
We evaluate our SVR model's performance on single-view reconstruction tasks and present both quantitative and qualitative results. We compare our method to two state-of-the-art 3D shape generation methods that use triangular mesh and implicit representation formats respectively: AtlasNet \cite{groueix2018} (with 1-Sphere and 25-Squares templates) and IM-NET \cite{Chen2019LearningIF}. Our results show that our method achieves significantly better results both in terms of reconstruction quality and fidelity without compromising from speed. We note that we omit comparing our method to voxel-based SVR methods, which suffer from much lower visual fidelity due to resolution constraints inherent to voxel-based generative methods. Finally, we show that our method enables reconstruction from real images with minimal preprocessing. 

\begin{table}
  \centering
  \newcommand{\qualimg}[1]{\raisebox{-.4\height}{\includegraphics[width=0.3\linewidth]{#1}}}
  \begin{tabular}{m{.5cm}cccc}
  Input  &
  \qualimg{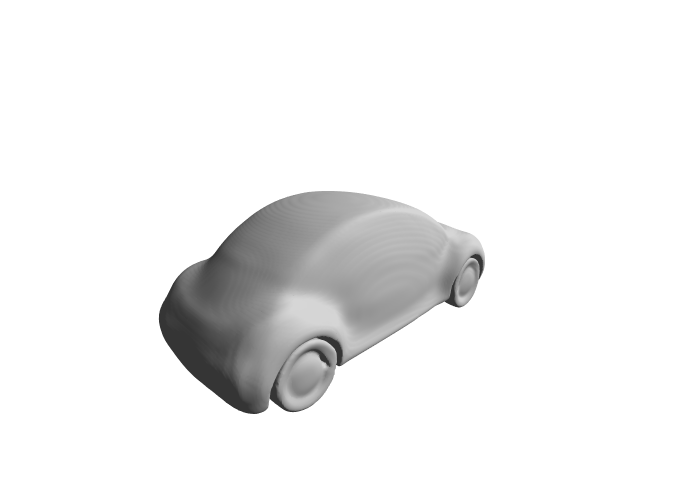} & 
  \qualimg{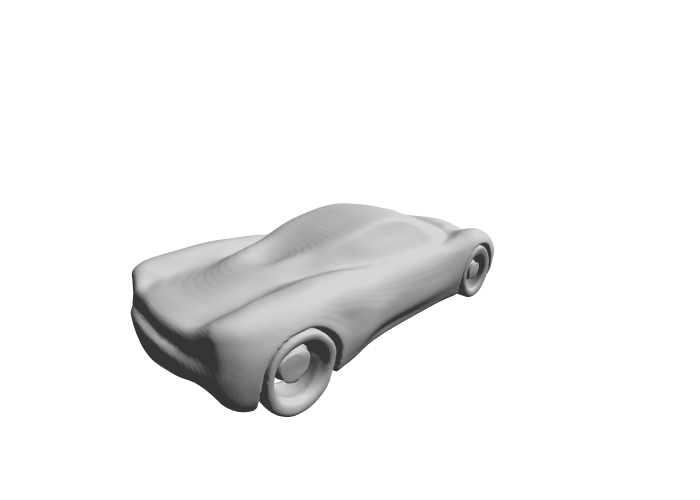}  \\
  \hline
 Result &
  \qualimg{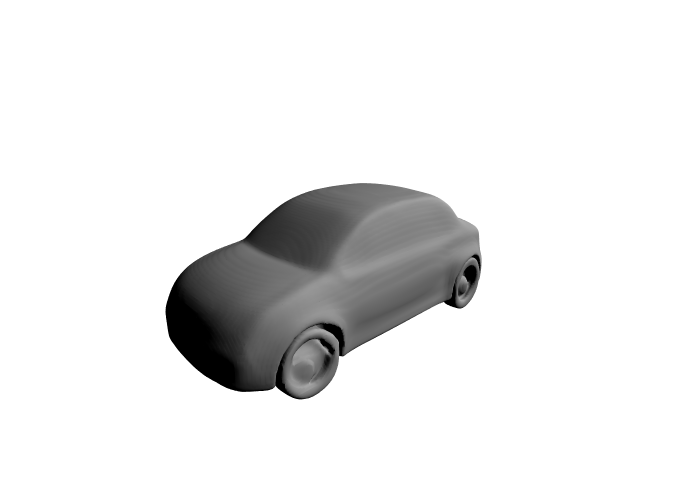} & 
  \qualimg{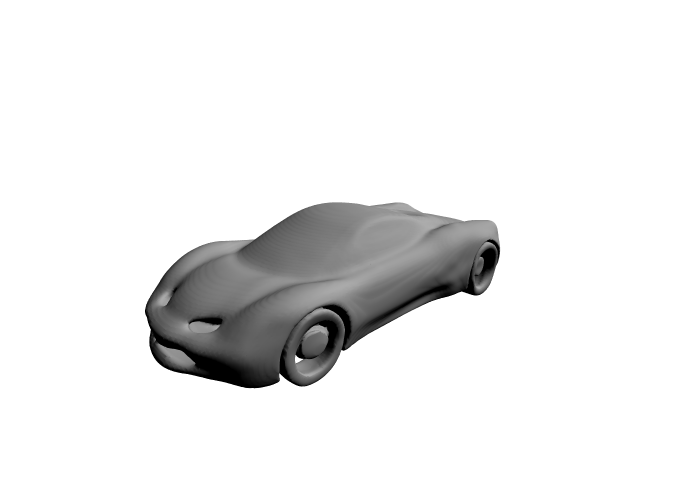} \\
  Input  &
  \qualimg{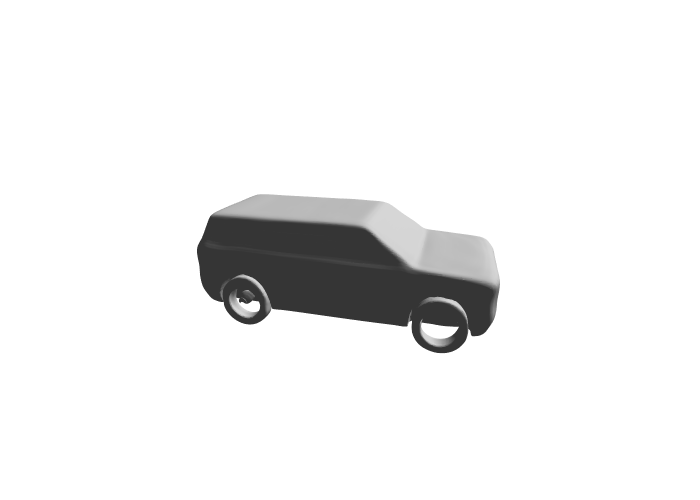} &  
  \qualimg{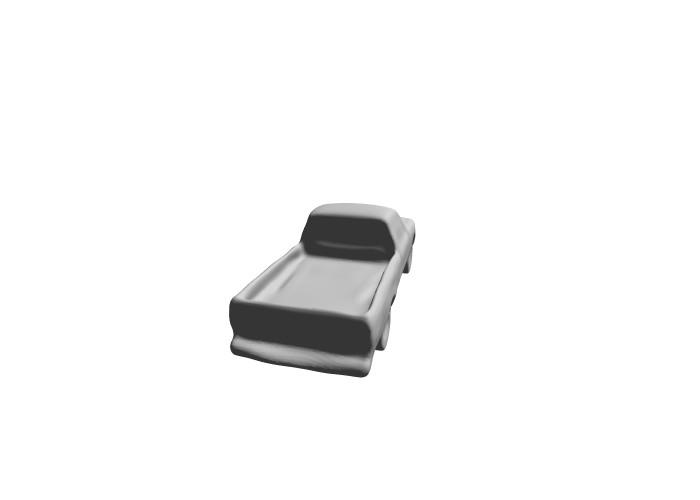}  \\
  \hline
 Result &
  \qualimg{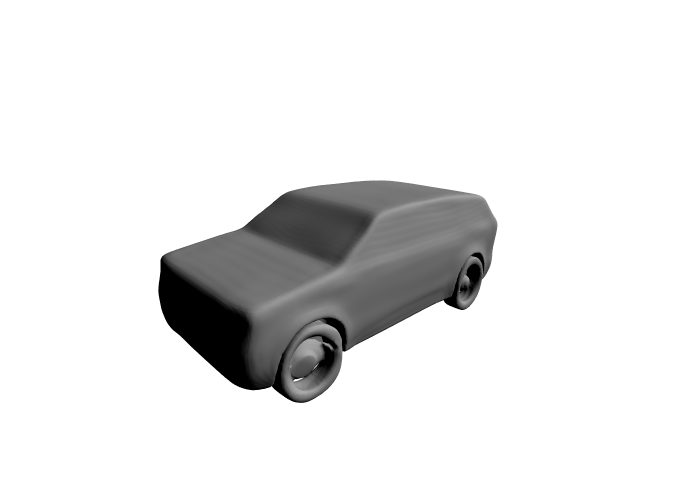} & 
  \qualimg{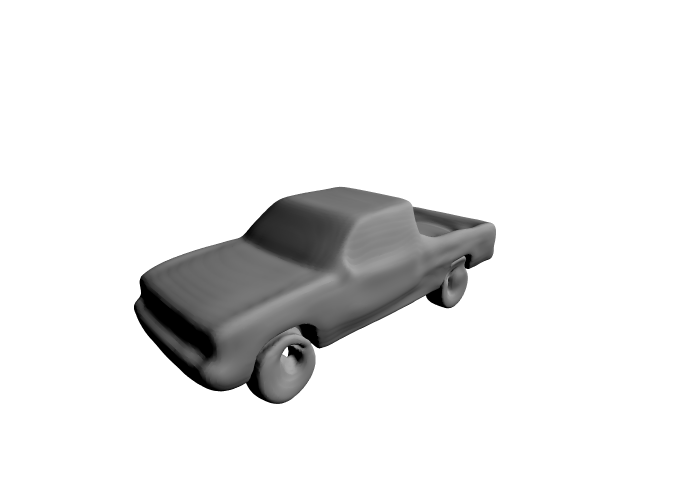} \\
  Input  &
  \qualimg{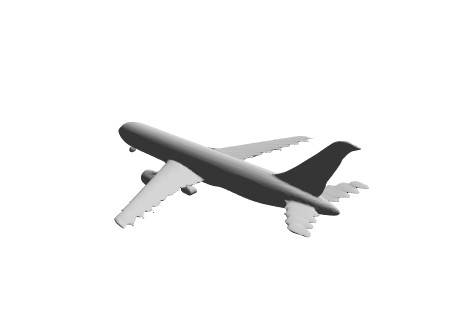} & 
  \qualimg{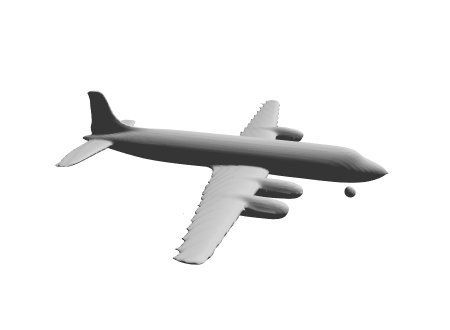}  \\
  \hline
 Result &
  \qualimg{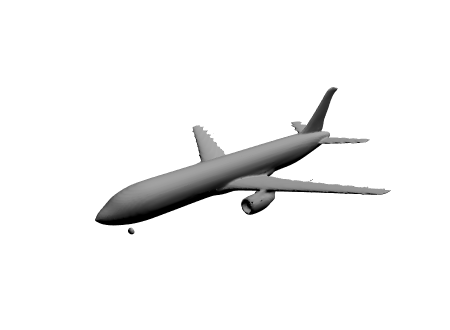} & 
  \qualimg{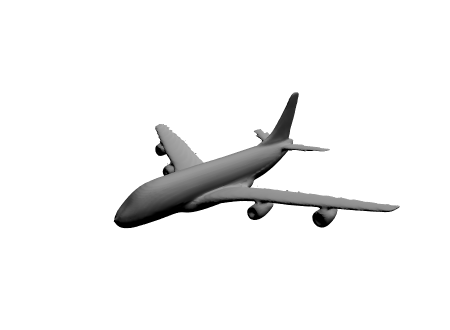} \\
  Input  &
  \qualimg{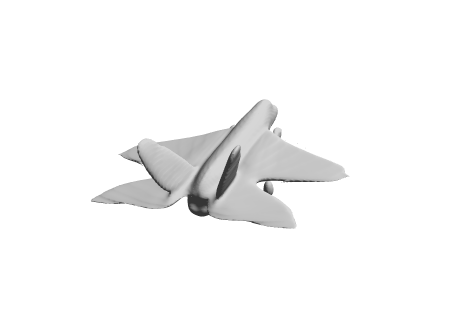} &  
  \qualimg{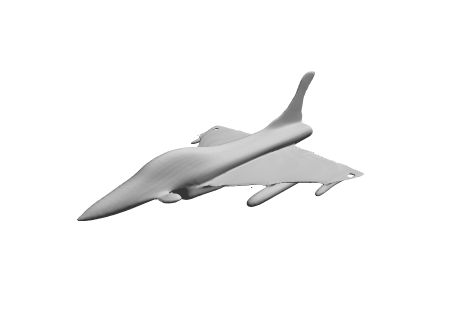}  \\
  \hline
 Result &
  \qualimg{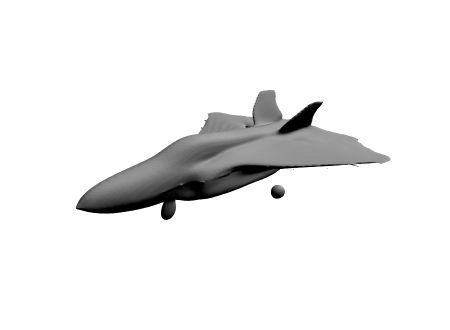} & 
  \qualimg{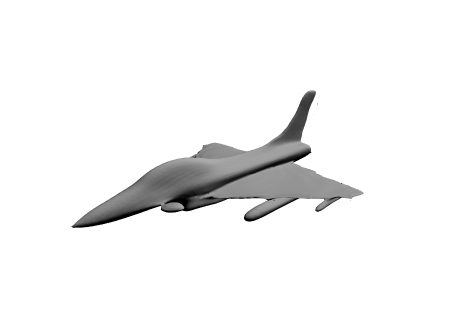} \\
  \end{tabular}
  \caption{Qualitative reconstruction results on cars: our method can successfully reconstruct 3D shapes from a variety of views.}
  \label{tb:qual_svr}
  \vspace{-1em}
\end{table}

We start with performing single-view reconstruction of various synthetic car and plane  images with our framework and present qualitative reconstruction results in Table \ref{tb:qual_svr}. As shown in Table \ref{tb:qual_svr}, our method not only reconstructs cars with high fidelity but can also infer unseen features such how the front of a sports car should look like. Similarly, a seemingly uninformative input image that depicts a truck's trunk is sufficient to reconstruct a truck. We also include cases where training set includes no similar samples, such as the Tortoise Beetle car depicted on the far left. In this case, we see that our method reconstructs a similar car that is able to capture the high-level semantics of the input image such as the overall size of the car, roundness of the roof and hood and, to a certain extent, the profile curves around the wheels. While the results on planes produce slightly more artifacts, Table \ref{tb:qual_svr} shows that our method can successfully reconstruct a variety of airplanes from different views. Note how our method is able to capture fine details such as the number of the engines, as well as the artifacts present in the input synthetic image.

Furthermore, we present a comparative qualitative analysis AtlasNet (with 1-Sphere and 25-Squares templates) and IM-NET using fixed-angle images and present the results in Table \ref{tb:qual_svr2}. As shown in Table \ref{tb:qual_svr2}, our method's performance significantly surpasses both AtlasNet 1-Sphere and 25-Squares methods and produces finer and accurate surface details compared to IM-NET. We note that while IM-NET's performance is close to our method in terms of capturing the overall shape, it produces more artifacts. Moreover, we note that IM-NET is not view-agnostic and is trained on fixed-view synthetic images. 

We also notice that even when our method cannot accurately reconstruct details such as the prominent front bumper of the car in the first input image, it successfully captures the overall design and style of the car such as the characteristic sharp profile curves. Similarly, we see that our method can successfully reconstruct finer details that the competitor methods cannot capture, such as the visible front landing gear seen in the commercial airplane image input. Interestingly, we see in the third sample of a fighter plane that our method introduces torpedoes to the reconstruction, a common feature within the fighter plane sub-category but not present in the input image. Hence, we deduce that our trained framework acts as a conceptual search engine due to using text-conditioned image features and can map the input image to the closest semantic point within the latent space. 

We next perform a quantitative analysis to measure the reconstruction fidelity of our method and competitor methods. To this end, we use the ground-truth meshes that correspond to the input synthetic images and compute the Earth Mover's Distance(EMD) and the Chamfer Distance between each reconstructed shape and corresponding ground truth, and report the average Chamfer Distance and EMD values in Table \ref{tb:quant-svr}.

\begin{table}[h]
\centering
\begingroup
\setlength{\tabcolsep}{5pt} % Default value: 6pt
\renewcommand{\arraystretch}{1.5}

\begin{tabular}{@{}|l|c|c|c|c|@{}}
\Xhline{2.0pt}
 & \multicolumn{4}{c|}{\textbf{SVR error} } \\
 \cline{2-5}
 & \multicolumn{2}{c|}{Chamfer $\downarrow$} 
 & \multicolumn{2}{c|}{EM Dist. $\downarrow$}  \\
 \cline{2-5} 
 & Car & Airplane & Car & Airplane  \\
\Xhline{2.0pt}
AtlasNet-Sph & 92.48  & 46.17  & 21.03  & 14.32  \\
AtlasNet-25 & 82.34  & 38.45  & 18.23  & 13.29  \\
IM-Net & 3.539  & 4.166  & 3.24  & 3.04  \\  
Ours & \bf{2.438}  & \bf{3.144}  & \bf{2.62}  & \bf{2.16} \\
\Xhline{1pt}
\end{tabular}
\endgroup
\vspace{0.5em}
\caption{Quantitative results on single-view reconstruction. Note that the CD values are multiplied by $10^{3}$ and EMD values are multiplied by $10^{2}$.}
\label{tb:quant-svr}
\vspace{-0.5em}
\end{table}

Results in Table \ref{tb:quant-svr} show that our method achieves the best CD and EMD values on both the car and airplane categories. We note that this is especially significant as both AtlasNet and IM-NET models are trained on fixed-view images for a fair comparison whereas our method is view-agnostic and trained on multi-view images. Given that airplanes often have many sharp and spiky features, view-agnostic SVR of planes is a more difficult task compared to cars. These results also demonstrate that our method is qualitatively and quantitatively superior to competitor methods.

\begin{table}[h!] 
\centering
\begingroup
\newcommand{\methodrow}{ & }%
\small{
\setlength{\tabcolsep}{2pt} % Default value: 6pt
\begin{tabular}{@{}|l|c|c|c|@{}}
\Xhline{2.0pt}
 & \textbf{Ours}  & \textbf{AtlasNet}  &  \textbf{IM-NET} \\
\Xhline{2pt}
Real-time generation   & No    & Yes    & No  \\
Real-time manipulation    & \textbf{Yes} & Yes & No  \\
High-quality surface    & \textbf{Yes} & No & Yes \\
High-quality reconstruction & \textbf{Yes} & No & Yes  \\
View-agnostic reconstruction & \textbf{Yes} & No & No  \\
\Xhline{2pt}
\end{tabular}
}
\endgroup
\vspace{0.5em}
\caption{A comparison of our framework with competitor methods. We include quantitative and qualitative comparisons with AtlasNet and IM-NET methods in our experiments. }
\label{tb:methods}
\vspace{-0.5em}
\end{table}

In addition to quantitative and qualitative analysis, we measure and report the run times of our SVR framework and all competitor methods, and present the results in Table \ref{tb:inf-time}. For a fair comparison, we take both inference time to predict the latent vector and the time required for ray-marching to construct a mesh into account. As shown in Table \ref{tb:inf-time}, our method is significantly faster than IM-NET while achieving better reconstruction results. While AtlasNet is much faster than our method due to directly outputting meshes, we note that our method is vastly superior to AtlasNet's performance in terms of reconstruction fidelity and quality as shown previously. 

\begin{table} [h]
\centering
\begingroup
\setlength{\tabcolsep}{5pt} % Default value: 6pt
\renewcommand{\arraystretch}{1.5}
\begin{tabular}{@{}|l|c|@{}}
\Xhline{2.0pt}
 & \multicolumn{1}{c|}{\textbf{Inference Time (s)} }\\
\Xhline{2.0pt}
AtlasNet-Sph & 0.078 \\
AtlasNet-25 & 0.112 \\
IM-Net & 13.280 \\
Ours & 8.030 \\
\Xhline{1pt}
\end{tabular}
\endgroup
\vspace{0.5em}
\caption{Comparison of inference time of our method and all competitor methods.}
\label{tb:inf-time}
\end{table}

Finally, we perform SVR using unseen real images, a notoriously difficult task. To this end, we first filter out the real image dataset of cars as described in Section \ref{sec:real-to-synth} and perform background removal using a pre-trained DETR \cite{Carion2020EndtoEndOD} segmentation model trained on the COCO 2017 Panoptic \cite{Lin2014MicrosoftCC} dataset. We then use our trained Real-to-Synthetic image translation model and translate real images to synthetic images. We use the output synthetic image as input to our 3D-LatentMapper framework to predict a latent vector and reconstruct the final mesh via ray-marching. We present the input images, predicted synthetic versions and reconstruction results of sample cars in Table \ref{tb:real-svr}.

As shown in Table \ref{tb:real-svr}, while our real-to-synthetic image translation model does not produce plausible synthetic images, it successfully learns to remove clutter and change the color scale of the target object (cars) to resemble that of our synthetic image dataset. Furthermore, despite the shortcomings of the image translation module, we find that our 3D-LatentFramework can still capture important, definitive attributes of the input images. Notice how in the first sample, the reconstructed car resembles both a Sedan and Hatchback car. Similarly, the reconstruction of the second sample, which is an Aston Martin, features characteristic Aston Martin attributes such as the front hood curves. However, the reconstruction of the fourth sample is not nearly as successful and does not feature any of the prominent attributes in the input image. We note that this is partly due to reconstructing from unseen views and partly due to rather low-quality image translation.

\begin{table}[t]
  \centering
  \newcommand{\qualimg}[1]{\raisebox{-.5\height}{\includegraphics[width=0.16\linewidth]{#1}}}
  \begin{tabular}{m{1cm}cccccc}
  Input  &
  \qualimg{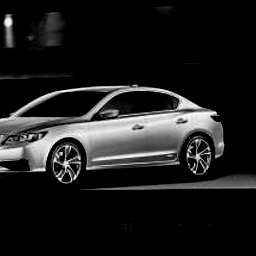} & 
  \qualimg{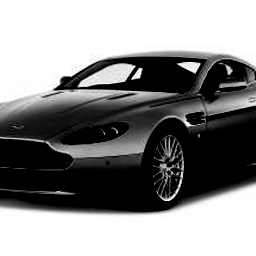} & 
  \qualimg{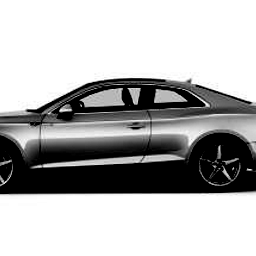} &  
  \qualimg{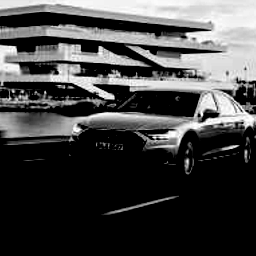}  \\
  \hline
 Synthetic &
  \qualimg{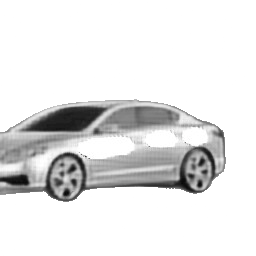} & 
  \qualimg{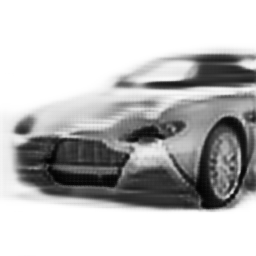} & 
  \qualimg{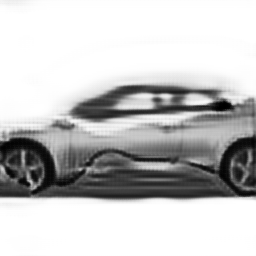} & 
  \qualimg{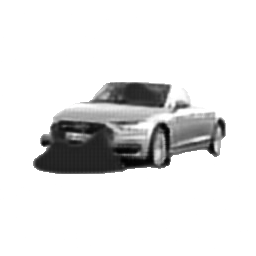} \\
 Output &
  \qualimg{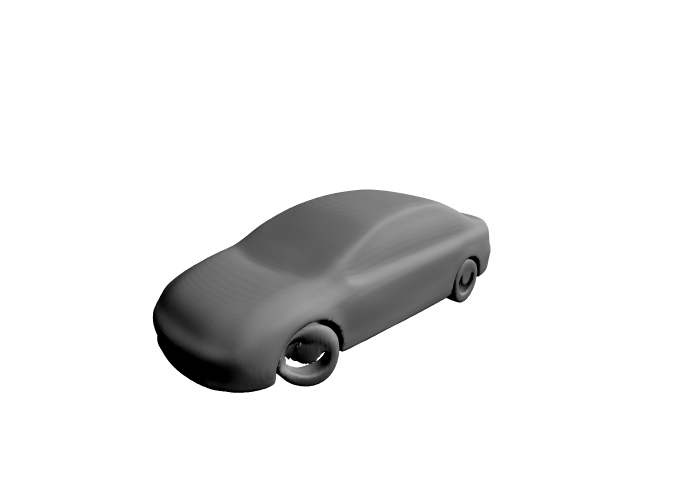} & 
  \qualimg{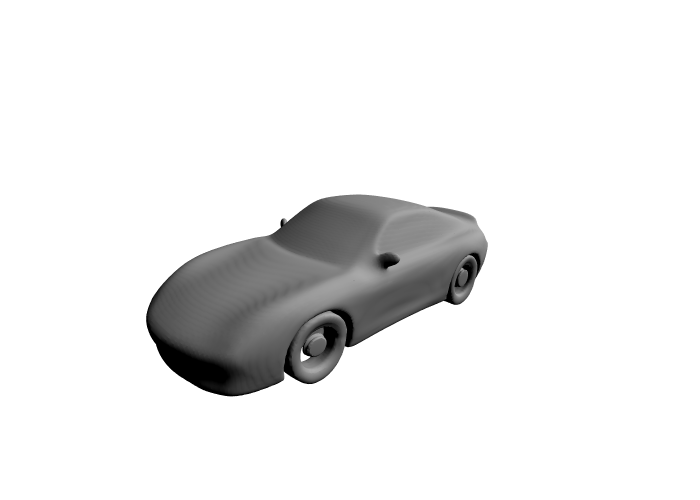} & 
  \qualimg{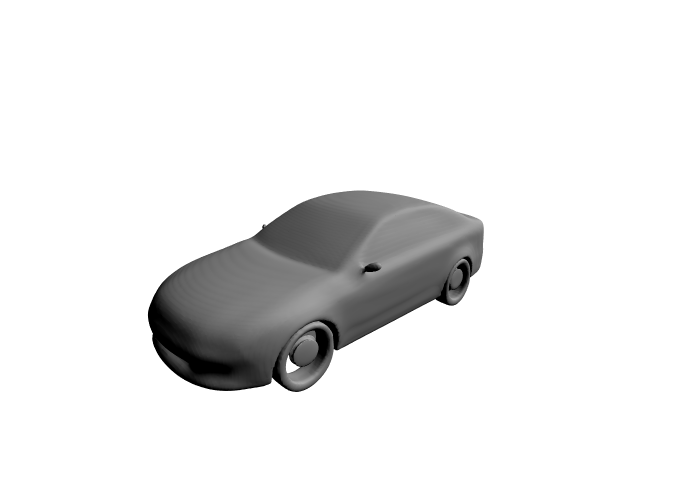} & 
  \qualimg{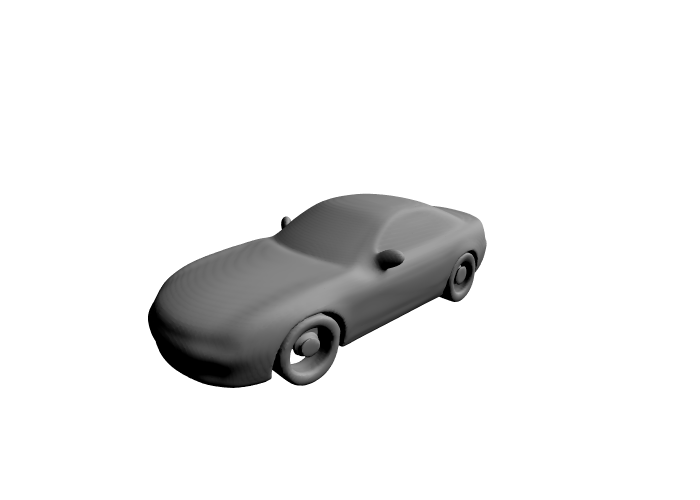} \\
 
  \end{tabular}
  \caption{Reconstruction from real images: we first perform real-to-synthetic image translation and use the predicted synthetic images as input to our 3D-LatentMapper framework.}
  \label{tb:real-svr}
  \vspace{-1em}
\end{table}

\section{Conclusion}
\label{chapter:conclusion}
In this work, we proposed 3D-LatentMapper, a view-agnostic single-view 3D shape reconstruction framework that directly operates within the learned latent space of a base 3D generative model, DIT. Through extensive experiments, we showed that our method can effectively map images into the latent space of DIT without pose constraints while outperforming state-of-the-art models both qualitatively and quantitatively. To the best of our knowledge, our method is the first that leverages deep positional and multi-modal features for Single-View Reconstruction. Our key takeaways are as follows:
\begin{itemize}
\item CLIP provides text-aligned dense features that are immensely increases the generalization capabilities of the SVR model to previously unseen objects.
\item Latent spaces of ViT are rich and expressive. We found that key descriptors extracted from the intermediate layers of ViT act as highly effective dense descriptors that are robust to changes in pose and scale.
\item Our proposed method is limited by the capabilities and expressiveness of the base 3D generative model's latent space. Hence, we believe a base model with a richer latent space would lead to better results.
\end{itemize}

{\small
\bibliographystyle{ieee_fullname}
\bibliography{egbib}
}

\appendix

\section{Network Architecture} 
\subsection{Deep ViT Autoencoder Architecture} 
\label{app:deep-vit}
We use an autoencoder architecture with feed forward layers to further compress extracted ViT descriptors. Our proposed network consists of 2 dense layers followed by ReLU activations. Once the model is trained, we use the frozen model and extract dense features from the bottleneck (output of the encoder). 

\subsection{Latent Mapper Architecture}
\label{app:latent-mapper}
We employ a dense neural network to map dense embeddings extracted from ViT and CLIP to the latent space of DIT. Our proposed network consists of 3 dense layers followed by hyberbolic tangent activations.

\section{Single View Reconstruction Experiment}
\label{app:svr-comp}
For SVR with AtlasNet \cite{groueix2018} and IM-NET \cite{Chen2019LearningIF}, we use the official implementations and the renderings of ShapeNet data provided by 3D-R2N2 \cite{choy20163d} with a 90-10\% split for training and testing. For AtlasNet experiments, we trained image encoders on \textit{car} and \textit{airplane} categories and used the decoder from the respective autoencoder with fixed parameters to compute a Chamfer distance loss between the resulting mesh and the ground truth mesh corresponding to the input image. For IM-NET experiments, we trained ResNet encoders on \textit{car} and \textit{airplane} categories and used the trained implicit decoder with fixed parameters to train a mapping network that maps images to the latent space of IM-NET. More specifically, we used grayscale images as input and trained ResNET encoders to minimize the mean squared loss between the predicted feature vectors and the ground truth feature vectors encoded by the pre-trained autoencoder.

\section{Sentence Templates for Prompt Engineering}
\label{app:templates}
We leverage CLIP to filter out irrelevant images in our real image datasets. Our method uses 74 sentence templates to compute the average CLIP distance between the target category and the content of each image. The list of templates we use for augmentation can be found in Table \ref{tab:templates-prompt}1. 

\setcounter{table}{0}
\renewcommand{\thetable}{C\arabic{table}}

\begin{table} [!t]
\begin{center}
\begin{tabular}{p{0.4\linewidth}p{0.4\linewidth}}
\hline `a bad photo of a' & `a sculpture of a' \\
\hline `a photo of the hard to see' &  `a low resolution photo of the' \\ 
\hline `a rendering of a' & `graffiti of a' \\
\hline `a bad photo of the' &  `a cropped photo of the' \\
\hline `a photo of a hard to see' & `a bright photo of a' \\
\hline `a photo of a clean' & `a photo of a dirty' \\
\hline `a dark photo of the' & `a drawing of a' \\ 
\hline `a photo of my' & `the plastic' \\ 
\hline `a photo of the cool' &  `a close-up photo of a' \\
\hline `a painting of the' & `a painting of a' \\
\hline `a pixelated photo of the' & `a sculpture of the' \\
\hline `a bright photo of the' & `a cropped photo of a' \\ 
\hline `a plastic' &  `a photo of the dirty' \\
\hline `a blurry photo of the' & `a photo of the' \\
\hline `a good photo of the' & `a rendering of the' \\
\hline `a in a video game.' & `a photo of one' \\
\hline `a doodle of a' &  `a close-up photo of the' \\
\hline `a photo of a' &  `the in a video game.' \\
\hline `a sketch of a' &  `a face of the' \\
\hline `a doodle of the' & `a low resolution photo of a' \\
\hline `the toy' & `a rendition of the' \\
\hline `a photo of the clean' &`a photo of a large' \\
\hline `a rendition of a' & `a photo of a nice' \\
\hline `a photo of a weird' & `a blurry photo of a' \\
\hline `a cartoon' &`art of a' \\
\hline `a sketch of the' & `a pixelated photo of a' \\
\hline `itap of the' & `a good photo of a' \\
\hline `a plushie' & `a photo of the nice' \\
\hline `a photo of the small' &`a photo of the weird' \\
\hline 'the cartoon' &`art of the' \\
\hline `a drawing of the' & `a photo of the large' \\
\hline `the plushie' & `a dark photo of a' \\
\hline `itap of a' & `graffiti of the' \\
\hline `a toy' & `itap of my' \\
\hline `a photo of a cool' & `a photo of a small' \\
\hline `a 3d object of the' &`a 3d object of a' \\
\hline `a 3d face of a' & `a 3d face of the' \\
\hline \end{tabular}
\vspace{1em}
\caption{List of templates that our method uses for augmentation. The input text prompt is added to the end of each sentence template.}
\end{center}
\label{tab:templates-prompt}
\end{table}

\end{document}